%% file: template.tex
\documentclass{Interspeech2024}
\usepackage{booktabs, multirow} % for borders and merged ranges
\usepackage{soul}% for underlines
\usepackage[table]{xcolor} % for cell colors
\usepackage{changepage,threeparttable} % for wide tables

\usepackage{amsmath,tabularx}
\usepackage{makecell}
\usepackage{tablefootnote}
\usepackage{changepage,threeparttable}

\usepackage{hyperref}
\usepackage{times}
\usepackage{textcomp}
\usepackage{cite}
\DeclareUrlCommand\website{\urlstyle{same}}

\interspeechcameraready

\title{DiscreteSLU: A Large Language Model with Self-Supervised \\ Discrete Speech Units for Spoken Language Understanding \vspace{-0.3cm}}

\name[affiliation={1}]{Suwon}{Shon}
\name[affiliation={1}]{Kwangyoun}{Kim}
\name[affiliation={1}]{Yi-Te}{Hsu}
\name[affiliation={1}]{Prashant}{Sridhar}
\name[affiliation={2}]{Shinji}{Watanabe}
\name[affiliation={3}]{Karen}{Livescu}

\address{
  $^1$ASAPP, USA \
  $^2$Carnegie Mellon University, USA  \
  $^3$Toyota Technological Institute at Chicago, USA}
\email{sshon@asapp.com \vspace{-0.5cm}}


\begin{document}

\maketitle


\begin{abstract}
The integration of pre-trained text-based large language models (LLM) with speech input has enabled instruction-following capabilities for diverse speech tasks. This integration requires the use of a speech encoder, a speech adapter, and an LLM, trained on diverse tasks. 
We propose the use of discrete speech units (DSU), rather than continuous-valued speech encoder outputs, that are converted to the LLM token embedding space using the speech adapter. We generate DSU using a self-supervised speech encoder followed by k-means clustering. 
The proposed model shows robust performance on speech inputs from seen/unseen domains and instruction-following capability in spoken question answering.
We also explore various types of DSU extracted from different layers of the self-supervised speech encoder, as well as Mel frequency Cepstral Coefficients (MFCC). Our findings suggest that the ASR task and datasets are not crucial in instruction-tuning for spoken question answering tasks.

% Our method is reproducible, using open models and publicly available datasets without any additional data generation methods such as text-to-speech generation or LLM-assisted label generation.

\end{abstract}

\section{Introduction}
\label{sec:intro}
\input{sections/introduction}

\input{figures/highlevel}

\section{Related work}
\label{sec:related}
\input{sections/related}
\input{tables/instruction_output}

\input{tables/exp_asr}

\input{tables/exp_slu}
\input{tables/exp_s2tt}

\section{Proposed approach}
\label{sec:proposed}
\input{sections/proposed}

\section{Experiments}
\input{tables/exp_dsu}
\label{sec:experiments}
\input{sections/experiments}

\section{Conclusion}
\label{sec:conclusion}
\input{sections/conclusion}

% \ifinterspeechfinal
%      The Interspeech 2024 organisers
% \else
%      The authors
% \fi
% would like to thank ISCA and the organising committees of past Interspeech conferences for their help and for kindly providing the previous version of this template.

\clearpage
\vspace{-0.1cm}

\bibliographystyle{IEEEtran}
\bibliography{mybib}

\end{document}

%% file: sections/introduction.tex
Recent work integrating pre-trained text-based large language models (LLMs) with speech input has enabled instruction-following capabilities for diverse speech processing tasks~\cite{pan2023cosmic,wang2023slm,tang2023salmonn,fathullah2023towards,chu2023qwen}. To feed speech into an LLM, additional modules consisting of a speech encoder and speech adapter are typically used. The speech encoder converts audio into a speech embedding sequence. Then, the speech adapter maps the speech embedding sequence into the text token embedding space of the LLM.  These previous studies typically use a speech encoder trained for automatic speech recognition (ASR) using human transcriptions. The length of the speech encoder output is then reduced via one of several compression approaches. Finally, the speech adapter and LLM are fine-tuned on a speech dataset with appropriate instructions, while the speech encoder is typically frozen.

%While these approaches have shown promising results, the speech encoder design has not been thoroughly investigated compared to the other parts. For example, self-supervised speech models exploiting a much larger set of speech data may generate better speech representations than ASR-supervised speech encoders. Moreover, a recent study on discrete units with length reduction methods shows notable performance in speech tasks which would be beneficial for feeding speech into LLM.

However, using self-supervised learning (SSL) speech models~\cite{hsu2021hubert,chen2022wavlm}, which can exploit a much larger set of speech data, may generate better speech representations than using single-task supervised %speech 
encoders.
% ~\cite{pan2023cosmic,wang2023slm,tang2023salmonn,fathullah2023towards,chu2023qwen}. 
Moreover, a recent study on discrete units~\cite{chang2023exploring} with length reduction%method
~\cite{wu2023wav2seq,lakhotia2021generative} shows promising performance across speech tasks, compared to using continuous SSL representations, and may be a good compromise between performance and efficiency.
%notable performance in speech tasks to compromise between its advantages and performance compared to using SSL speech representation as it is. 
We believe that this approach could also be beneficial for feeding speech into an LLM. 
% However, DSU still needs SSL speech representation from a deep layer to ensure robust performance. 

This paper studies discrete speech units (DSU) combined with a speech adapter to convert 
%discrete units 
DSU into the embedding space of LLM tokens. The DSU can be extracted using any layer of an SSL speech model, and also can be extracted using acoustic features such as MFCCs to significantly reduce the computation load.
The key contributions of this work are as follows:

\begin{itemize}[leftmargin=0.5cm]
    \item We compare ASR-trained speech encoders and DSU-based speech input, with quantitative analyses on seen/unseen speech domains and zero-shot tasks.
    \item We investigate various types of DSUs as inputs to large language models, ranging from deep to shallow layers and even MFCCs. 
    \item We present a reproducible method for building our models, using publicly available datasets without any additional data mining methods, text-to-speech data generation, or LLM-based data generation.
\end{itemize}

%% file: figures/highlevel.tex
\begin{figure}[t]
    \centering
    \includegraphics[width=0.5\textwidth]{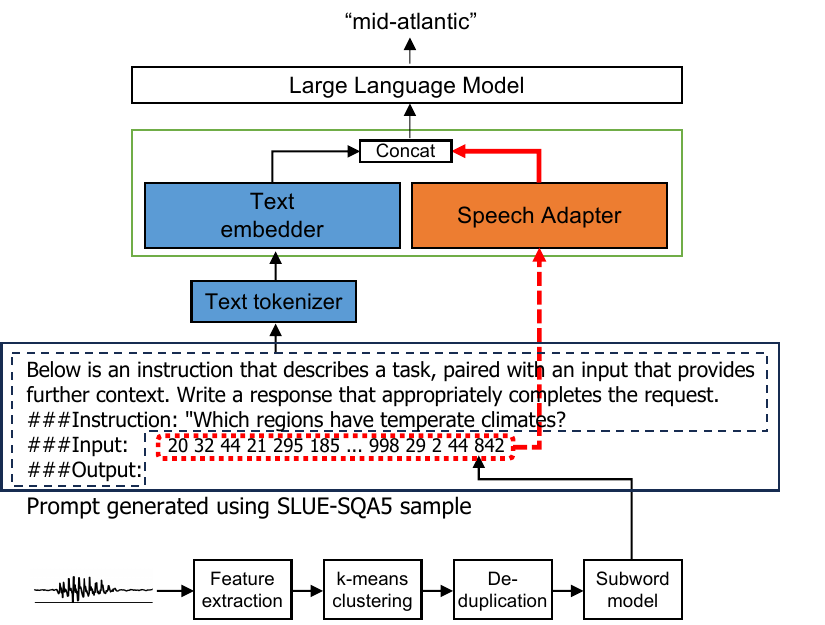}
    % \vspace{-0.5cm}
    \caption{Model architecture overview}
    \label{fig:highlevel}
    \vspace{-0.7cm}
\end{figure}

%% file: sections/related.tex
\vspace{-0.1cm}

There has been growing interest in, and a number of innovative approaches for, using LLMs to enable traditional speech tasks such as ASR and SLU via natural language prompts or instructions. 
The most common approach is to use a speech encoder trained for ASR using either a connectionist temporal classification (CTC)~\cite{lakomkin2023end,wu2023decoder,pan2023cosmic,fathullah2023prompting,fathullah2023towards} or an attention-based encoder-decoder (AED)~\cite{wang2023slm,tang2023salmonn,yu2023connecting,chu2023qwen} approach. For CTC, length compression is achieved by removing embedding frames corresponding to the blank token or by averaging embeddings that belong to consecutive %unique
same-label tokens~\cite{pan2023cosmic,wu2023decoder,tsunoo2023decoder}. Down-sampling with striding can also be applied in both approaches~\cite{lakomkin2023end,fathullah2023prompting,fathullah2023towards,hono2023integration,wang2023slm,yu2023connecting}.

These approaches have shown competitive speech task performance with instruction following capability and even zero-shot or 1-shot capability~\cite{pan2023cosmic,wang2023slm,chen2023salm,gong2023joint}. However, some studies train and evaluate on fixed tasks such as ASR or speech-to-text translation (S2TT)~\cite{wu2023decoder,fathullah2023prompting,lakomkin2023end,yu2023connecting,hono2023integration,chen2023salm}; these models do not follow instructions and perform only the limited tasks they have seen during training. Furthermore, most of the prior work lacks out-of-domain evaluation or has not been thoroughly benchmarked.%~\cite{chen2023salm}.
% Instruction following with speech input has been successfully shown in several models~\cite{pan2023cosmic,wang2023slm,tang2023salmonn,fathullah2023towards,chu2023qwen} and even zero shot or 1 shot capability was shown in~\cite{pan2023cosmic,wang2023slm}. However, some studies train and evaluate only on fixed tasks like ASR or speech-to-text translation (S2TT)~\cite{wu2023decoder,fathullah2023prompting,lakomkin2023end,yu2023connecting,hono2023integration}; these models do not follow instructions and perform only the limited tasks they have seen during training.
% These approaches have shown competitive speech task performance on domains matched to the training domain, but there is a lack of out-of-domain evaluation. 
%This paper intensively
In this work, we extensively investigate the out-of-domain capability of the proposed method. %to check if the model could be over-fitted to a specific domain by the type of speech encoder used.

%In addition, many studies generate these datasets automatically, 
Some prior work uses training datasets that are generated automatically via LLM-assisted text label generation, using models such as GPT 3.5~\cite{pan2023cosmic,tang2023salmonn} or LLama-2\cite{fathullah2023towards}, and audio generation via TTS~\cite{wang2023slm,fathullah2023towards}.
The evaluation is also often done using the same data generation process \cite{pan2023cosmic,fathullah2023towards}, which may not reflect the real-world performance of the model on the task. In this study, we use enitrely open-source datasets for training, without any TTS-generated audio or LLM-generated labels, and the evaluation is done with human labels.

While many studies investigate discrete units for speech~\cite{chou2023toward,nguyen2024spirit,lakhotia2021generative}, the most relevant prior approach to ours is AudioPaLM~\cite{rubenstein2023audiopalm}.
Rather than using a speech adapter, AudioPaLM feeds discrete audio tokens into the model directly to generate audio/text. However, the tasks in this prior work are limited to ASR and translation; the model's ability to follow instructions or perform other SLU tasks has not yet been evaluated. For extracting the discrete audio tokens, AudioPaLM considered multiple speech representation models such as w2v-BERT~\cite{chung2021w2v} and USM~\cite{zhang2023google}. However, the study did not consider the effect of extracting embeddings and discrete units from different layers in the same model.  As shown in prior work~\cite{pasad2021layer}, different speech attributes are encoded in different model layers, so the choice of layer may  impact considerably both downstream performance and computational load. 
VoxtLM~\cite{maiti2023voxtlm} also considers DSUs as input to a language model for multitask learning,
but using task tokens 
%to perform one of the multitask learned during training 
rather than instruction-following.
%with task token but with no instruction following ability.
Finally, SpeechGPT~\cite{zhang2023speechgpt} also uses DSUs
%discrete speech units 
as input to an LLM, but
%. However, 
the evaluation is qualitative via %brief 
examples;
%and 
there is no quantitative performance evaluation.
In contrast to all these studies, we quantitatively evaluate instruction-following ability for spoken language understanding, explore DSUs extracted from different layers,
%that may contain different information for speech. 
%Furthermore, we evaluated on the 
and also evaluate DSUs extracted using MFCCs, which may significantly reduce both training effort and computational load.

%% file: tables/instruction_output.tex
\begin{table}[!t]\centering
\footnotesize
\caption{Instructions and outputs for our speech tasks.}\label{tab:task_prompt}
\vspace{-0.2cm}
\begin{tabular}{l|l}\toprule
% Task &Instruction &Output \\\midrule
SQA & Intruction: \{question\} \\
& Output: \{answer\} \\\midrule
ASR & \makecell{Instruction: Generate transcription of the given}\\ 
& \phantom{Instruction:} speech input \\ 
& Output: \{normalized transcrition\} \\ \midrule
SA & Instruction: Classify the given speech into one \\
&\phantom{Insrn:} of positive, neutral and negative sentiments\\
& Output: \{positive, neutral, negative\}\\ \midrule
NER: & Instruction: Find named entity in the speech. \\
& Output: set of \{phrase (type of named entity)\} \\ \midrule
S2TT: &Instruction: Translate the input to \{language\} \\
\bottomrule
\end{tabular}
\vspace{-0.5cm}
\end{table}

% \begin{table}[t]\centering
% \caption{Instruction and output example for speech tasks}\label{tab:task_prompt}
% \vspace{-0.2cm}
% \begin{tabularx}{\linewidth}{lll}\toprule
% Task &Instruction &Output \\\midrule
% SQA: &\{question\}& \{answer\} \\
% ASR: &Generate transcription of the given speech input& \{normalized transcrition\} \\
% SA: &Classify the given speech into one of positive, neutral and negative sentiments& \{positive, neutral, negative\} \\
% NER: &Find named entity in the speech. &set of \{phrase (type of named entity)\} \\
% S2TT: &Translate the input to {language} &- \\
% \bottomrule
% \end{tabularx}
% \end{table}

%% file: tables/exp_asr.tex
\begin{table*}[t]\centering
\caption{ASR task trained system evaluation. A and B are traditional ASR systems. C, D, E, and F use compressed CTC embeddings as input to the speech adapter+LLM. G, H and J are our systems that use DSUs as input to the speech adapter+LLM.}\label{tab:exp_asr}
\footnotesize
\vspace{-0.3cm}
\begin{tabular}{rll|crrr|c}\toprule
& \multirow{2}{*}{System ID} &System type&\multicolumn{4}{c|}{WER} &BLEU-1 \\ \cmidrule{4-8}
& & &\makecell{Tedlium3 \\ (dev/test)~\cite{hernandez2018ted}} &\makecell{SLUE-VP\\(test)~\cite{shon2022slue}} &\makecell{SLUE-VC\\(test)~\cite{shon2022slue}} &\makecell{Fleurs\cite{conneau2023fleurs}\\(test\_en\_us)} &\makecell{SLUE-SQA5 \cite{shon2022slue2} \\(dev / test / v-test)}\\ \midrule
A &Fbank-CTC-AED$^a$& Traditional ASR &\phantom{0}7.3 / \textbf{\phantom{0}6.4} &- &- &- & - \\
B &DSU(WavLM/21)-CTC-AED$^b$ & Traditional ASR &\phantom{0}9.0 / \phantom{0}8.9 &- &- &- & - \\ \hline
C &COSMIC-ASR-7B~\cite{pan2023cosmic}& CTC-LLM&\phantom{0.}-\phantom{0} / 13.1 &- &- &- & - \\
D &Fbank - CTC (blank-removed)$^a$ & CTC-LLM &21.1 / 16.8 &25.0 &32.4 &37.6 &1.2 / 1.1 / 1.5 \\
E &Fbank - CTC (averaged)$^a$$_{\textcolor{red}{baseline}}$ &CTC-LLM &10.1 / \phantom{0}8.4 &16.8 &24.5 &27.1 &1.4 / 1.4 / 1.9 \\
F &DSU(WavLM/21) - CTC(averaged)$^b$ &CTC-LLM &11.2 / \phantom{0}9.0 &12.7 &22.3 &14.4 &1.5 / 1.5 / 1.9 \\ \hline
G &DSU(WavLM/21) &DSU-LLM&\phantom{0}6.6 / \textbf{\phantom{0}6.4} &12.0 &\textbf{17.2} &13.0 &1.5 / 1.5 / 2.0 \\
H & \hspace{0.6cm}+ dedup &DSU-LLM &\phantom{0}8.1 / \textbf{\phantom{0}6.4} &11.8 &17.3 &\textbf{12.6} &1.5 / 1.5 / 2.0 \\
J & \hspace{0.6cm}+ dedup + subword &DSU-LLM&\phantom{0}7.7 / \phantom{0}7.1 &\textbf{11.2} &17.4 &13.2 &1.5 / 1.5 / 1.9 \\
\bottomrule
\end{tabular}
\\ \scriptsize{$^a$ctc model used here: \website{https://huggingface.co/espnet/dongwei_tedlium3_asr_e-branchformer_external_lm}}

\scriptsize{$^b$ctc model used here: \website{https://huggingface.co/espnet/kohei0209_ted3_asr2_e_branchformer1_raw_wavlm_large_21_km1000_bpe_rm2000_bpe_ts500_sp}}

\end{table*}

%% file: tables/exp_slu.tex
\begin{table*}[t]\centering
\vspace{-0.2cm}
\footnotesize
\caption{ASR + SLU task trained system evaluation. Note: the verified test set (v-test) for SLUE-SQA5 is a human-validated test set \cite{shon2022slue2}. $^*$: human transcription fine-tuned text-based model as an upper bound.}\label{tab:exp_slu}
\vspace{-0.3cm}
\begin{tabular}{lc|lccccc}\toprule
\multicolumn{2}{c}{\multirow{3}{*}{System ID}} & &\multicolumn{4}{c}{WER} & BLEU1 \\
& & Training data &\makecell{Tedlium3 \\ (test)} &\makecell{SLUE-VP\\(test)} &\makecell{SLUE-VC\\(test)} &\makecell{Fleurs\\(test\_en\_us)} &\makecell{SLUE-SQA5\\(dev/test/v-test)}\\ \midrule
E1 &\multirow{3}{*}{\makecell{CTC\\(averaged)\\$^{\textcolor{red}{baseline}}$}} &Tedlium 3 + SLUE-SQA5 & 8.5 &17.5 &24.9 &26.7 &62.1 / 56.4 / \textbf{50.5} \\ \cmidrule{3-8}
E2 & &\makecell{+ SLUE-VC\_\{ASR, SA\} \\+ SLUE-VP\_\{ASR, NER\}} & 9.7 &18.6 &33.5 &32.8 &57.9 / 52.4 / 46.6 \\ \midrule
G1 &\multirow{3}{*}{\makecell{DSU\\(WavLM/21)} }&Tedlium 3 + SLUE-SQA5 & 7.9 &12.1 &17.9 &13.9 &\textbf{62.7} / \textbf{56.8} / 48.2 \\ \cmidrule{3-8}
G2 & &\makecell{+ SLUE-VC\_\{ASR, SA\} \\+ SLUE-VP\_\{ASR, NER\}} & \textbf{6.8} &\textbf{11.2} &\textbf{15.3} &\textbf{13.0} &62.1 / 55.8 / 49.9 \\ \midrule
J1 & \multirow{4}{*}{\makecell{DSU\\(WavLM/21) \\ +dedup+subword}} &Tedlium 3 + SLUE-SQA5 & 8.5 &14.1 &19.9 &17.5 &60.7 / 55.3 / 48.3 \\ \cmidrule{3-8}
J2 & &\makecell{+ SLUE-VC\_\{ASR, SA\} \\+ SLUE-VP\_\{ASR, NER\}}&8.6 &13.0 &18.4 &15.3 &60.1 / 54.8 / 47.6 \\ \midrule
K1$^*$ & Mistral-7B-v0.1 &\makecell{SLUE-\{SQA5,VC\_SA,VP\_NER\}}&- &- &- &- &  83.2 / 78.8 / 77.2 \\
\bottomrule
\end{tabular}
\vspace{-0.3cm}
\end{table*}

%% file: tables/exp_s2tt.tex
\begin{table}[!t]\centering
\footnotesize
\caption{Zero-shot S2TT evaluation on Fleurs test set}\label{tab:exp_s2tt}
\vspace{-0.3cm}
\begin{tabular}{lcrrr}\toprule
\multicolumn{2}{c}{\multirow{2}{*}{System ID}} &\multicolumn{3}{c}{BLEU} \\\cmidrule{3-5}
& &En$\rightarrow$Fr &En$\rightarrow$De &En$\rightarrow$Es \\\midrule
K2 &\makecell{\makecell{Mistral-7B-\\Instruct-v0.1}} &19.01 &12.08 &16.18 \\\midrule
E1 &\multirow{3}{*}{\makecell{CTC (averaged)\\$^{\textcolor{red}{baseline}}$}} &0.11 &0.09 &0.10 \\
E2 & &6.71 &\textbf{3.09} &3.23 \\\midrule
J1 &\multirow{2}{*}{\makecell{DSU(WavLM/21)+\\dedup + subword}} &0.13 &0.13 &0.12 \\
J2 & &\textbf{8.52} &3.01 &\textbf{5.27} \\
\bottomrule
\end{tabular}
\vspace{-0.7cm}
\end{table}

%% file: sections/proposed.tex
% Figure~\ref{fig:highlevel} gives a high-level overview of our approach.

\subsection{Generating speech discrete units}
\vspace{-0.2cm}

Speech discretization is the process of transforming an audio waveform $X = [x_1,...,x_L]$, where  $x_l \in \mathbb{R}$, into a sequence of DSU $Z=[z_0,...,z_T]$, where $T\le L$ and $z_t \in \{1,...,K\}$. 
% DSU can be used for target labels for training self-supervised speech representation models that have shown robust performance on speech representation and are versatile enough to be fine-tuned for downstream tasks\cite{baevski2020effectiveness,chung2021w2v,hsu2021hubert}. On the other hand, DSU can be also used for input to the model. In this case, the system is inherently a two-step pipeline system, however, it has advantages in data storage and transmission compared to high dimensional embeddings and robust performance by employing an open-sourced large trained speech representation model for the downstream task with comparatively smaller size network~\cite{chang2023exploring,rubenstein2023audiopalm,lee2021direct}.
Our tokenization approach is based on k-means clustering of pre-trained self-supervised representations, similarly to previous studies~\cite{hsu2021hubert,chang2023exploring}. 
First, the speech input $X$ is mapped to a robust high-dimensional speech representation sequence $H = [h_0,h_1,...,h_T]$ using a pre-trained self-supervised model such as WavLM~\cite{chen2022wavlm} or HuBERT~\cite{hsu2021hubert}. 
We then use k-means clustering to convert the embedding sequence $H$ to a k-means cluster index sequence $Z$.
The resulting cluster index (DSU) sequence can be used directly as input to an LLM or can be further compressed by length reduction methods like de-duplication and subword modeling.
De-duplication involves converting sub-sequences of consecutive repeated cluster indices to a single unique index. 
A subword model can then be trained on the de-duplicated cluster index sequences to convert $Z$ to a meta-cluster index sequence $ \widetilde{Z} = [z_{0}, z_{1},..., z_{\widetilde{T}}]$ where $\widetilde{T} < T$ is the final reduced length in frames. 
The overall length reduction ratio $\widetilde{T}/T$ depends on the subword size and typically ranges between 30\% and 60\%.

\vspace{-0.2cm}

\subsection{Intruction-tuning with a speech adapter}
\vspace{-0.2cm}

For instruction tuning, we combine the DSU sequence $\widetilde{Z}$ with a text instruction and the desired output as shown in Figure~\ref{fig:highlevel}. 
% Although the DSU is compact and composed with text together, the DSU should be processed with the speech adapter separately. 
The speech adapter converts the DSU portion of the prompt into a continuous embedding sequence that aligns with the text token embedding space of the LLM.
The text portion of the prompt is tokenized into subwords, using the LLM's tokenizer, and is mapped to a continuous embedding sequence using the LLM's embedding lookup table. 
The two continuous embedding sequences, one from the DSU sequence and one from the text, are concatenated in the same order as in the original prompt.
Using this concatenated input sequence, along with the desired output text, instruction-tuning consists of updating the parameters of the speech adapter and LLM. 
For the LLM parameters, we use parameter-efficient fine-tuning, specifically Low-Rank Adaptation (LoRA)~\cite{hu2021lora}.

\vspace{-0.2cm}

\subsection{Instruction-tuning with a diverse dataset}
\vspace{-0.2cm}
A diverse instruction-tuning dataset is necessary to enable the instruction-following capability of the LLM.
One common way to build diverse (instruction, input, output) triplets is through LLM-assisted label generation, such as generating  translation labels given a monolingual text dataset. Speech input can be generated using a TTS system. However, such automatic dataset generation is a form of weak supervision, depends 
%and is dependent 
on the LLM and TTS performance, and requires potentially high inference costs for the large generation models.
In this study, we do not use any LLM-assisted label or TTS-generated speech.
All of our training data is based on SLUE~\cite{shon2022slue,shon2022slue2}, Tedlium 3~\cite{hernandez2018ted}, and Fleurs~\cite{conneau2023fleurs} to provide ASR, spoken question answering (SQA), sentiment analysis (SA), and named entity recognition (NER) examples for training.
For the SQA task, We use the provided question as the instruction. 
For other tasks, we use the simple instructions given in Table~\ref{tab:task_prompt}. 
We also include S2TT as a task only for evaluation; this task is unseen in training.

%% file: tables/exp_dsu.tex
\begin{table*}[!tp]\centering
\footnotesize
\caption{Comparison of DSUs extracted from shallow layers and MFCCs. *note that E3 is not DSU-based approach}\label{tab:exp_dsu}
\vspace{-0.2cm}
\begin{tabular}[t]{lr|c|rrrr|c}\toprule
\multicolumn{2}{c|}{\multirow{2}{*}{System ID}} &\textbf{} &\multicolumn{4}{c|}{WER} &BLEU1 \\
& & Training data &\makecell{Tedlium3 \\ (test)} &\makecell{SLUE-VP\\(test)} &\makecell{SLUE-VC\\(test)} &\makecell{Fleurs\\(test\_en\_us)} &\makecell{SLUE-SQA5\\(dev/test/v-test)}\\ \hline
J2 &WavLM/21 &\multirow{6}{*}{\makecell{Tedlium 3\\+ SLUE-SQA5\\ + SLUE-VC\_\{ASR, SA\} \\+ SLUE-VP\_\{ASR, NER\}}} & \textbf{8.6} &\textbf{13.0} &\textbf{18.4} &\textbf{15.3} &60.1 / 54.8 / 47.6 \\
J3 &WavLM/15 & &12.7 &19.1 &29.3 &25.8 &\textbf{61.7} / \textbf{56.8} / \textbf{50.9} \\
J4 &WavLM/10 & & 18.4 &32.3 &43.8 &37.7 &58.8 / 54.2 / 46.4 \\
J5 &WavLM/5 & & 40.7 &55.6 &76.8 &79.1 &55.3 / 51.5 / 44.9 \\
J6 &WavLM/1 & & 64.5 &92.4 &107.1 &147.3 &51.7 / 47.7 / 41.9 \\
L1 &MFCC & & 127.9 &113.3 &113.3 &162.2 &51.4 / 46.3 / 40.8 \\ \midrule
E3 &CTC (averaged)* &\multirow{3}{*}{\makecell{SLUE-SQA5 \\+ SLUE-VC\_SA \\+ SLUE-VP\_NER}} & 147.2 &105.6 &159.0 &126.7 &53.1 / 49.7 / 42.3 \\
J7 &WavLM/21 & & 151.6 &164.8 &126.6 &197.9 &\textbf{54.2} / 49.3 / 43.2 \\
L2 &MFCC & &\textbf{99.0} & \textbf{98.9} & \textbf{98.2} &\textbf{96.6} &53.2 / \textbf{50.0} / \textbf{43.3} \\
\bottomrule
\end{tabular}
\vspace{-0.4cm}
\end{table*}

%% file: sections/experiments.tex
\subsection{Model and training setup}
\vspace{-0.1cm}

To generate DSUs, we extract speech representations using WavLM and then cluster using k-means. We choose layer 21 following a previous study~\cite{chang2023exploring} that considered the CCA similarities with word labels~\cite{pasad2021layer}. 
% We analyze the impact of changing the chosen layer to one of \{1,5,10,15\}.
% We also consider MFCCs as an alternative to WavLM representations, using a hop size of 20ms and 50ms window to match the WavLM frame rate.
For k-means clustering and subword modeling, following~\cite{chang2023exploring} we use $K=1000$ and 2000 subword tokens.
% For k-means clustering, we use $K=1,000$ and train on a randomly sub-sampled 10\% of the Tedlium 3 train set. For subword modeling, we learn a vocabulary of 2,000 subword tokens.
For comparison, we reproduce CTC-based representation and compression approaches~\cite{pan2023cosmic}, including both blank removal and frame averaging.

For the speech adapter, an embedding layer first converts DSU to 512-dimensional embeddings. The rest of the adapter structure follows~\cite{pan2023cosmic}, with two 2D convolution layers with 2 stride, 4 transformer layers, and a linear layer. The linear layer converts the 512-dimensional transformer output to 4096 dimensions, to match the token embedding dimensionality of the LLM.
For the LLM component, we initialize the model weights with the Mistral-7B-v0.1 pre-trained model\cite{jiang2023mistral}. We use LoRA fine-tuning with a rank of 8 and $\alpha=$16 on all four projection layers in the transformer. The speech adapter has 18M trainable parameters, and the LoRA adapter has 7M parameters.

We conduct experiments on 8 A6000 GPUs, using an AdamW optimizer with minibatch size of 1280 for 15 epochs and 0.005 learning rate. The validation loss is evaluated every 200 steps.
\vspace{-0.1cm}

\subsection{Task and data specifications}
\vspace{-0.1cm}
We use various datasets for different tasks.
For ASR, we use the Tedlium 3~\cite{hernandez2018ted}, SLUE-VoxCeleb (SLUE-VC)~\cite{shon2022slue}, SLUE-VoxPopuli (SLUE-VP)~\cite{shon2022slue}, and Fleurs~\cite{conneau2023fleurs} datasets. For SQA, we use SLUE-SQA5~\cite{shon2022slue2}. We also use SLUE-VC for SA, SLUE-VP for NER, and Fleurs for S2TT.  Note that not all datasets are used for training in all experiments, depending on the experiment setup. For example, the Fleurs dataset is used only for evaluation in all experiments.
% Fleurs was not used at all for training in all experiments.

To validate model performance 
%from a diverse perspective, 
in diverse settings, we gradually add 
%SLU 
tasks starting from ASR-only, then SQA, and then SA and NER %with more 
plus additional ASR data. For ASR-only training, we use the Tedlium 3 dev set as a validation set. For all other experiments, we use the SLUE-SQA5 dev set as a validation set. 
Our evaluation metrics are word error rate (WER) for ASR and BLEU score for S2TT. For SQA, we use BLEU with a maximum n-gram order of 1 (i.e., we use BLEU-1) since %SLUE-SQA5 is closed-end SQA dataset and 
the majority of answers contain 1-2 words. Note that our SLUE-SQA5 task is slightly different from the original task in SLUE~\cite{shon2022slue2} since we modify the output for general QA as shown in Table~\ref{tab:task_prompt}.

\subsection{Results}
\vspace{-0.1cm}
\subsubsection{ASR-only task}
\vspace{-0.1cm}
%Although understanding and reasoning are considered more high-level speech tasks, speech recognition is still a fundamental task that a generalized speech model should achieve. In this experiment, w
Our first experiment focuses on the ASR task alone by training the model using the Tedlium 3 training set. The results are shown in Table~\ref{tab:exp_asr}. Similarly to previous work~\cite{chang2023exploration}, length reduction does not always give the best result, but 
%still it 
is efficient %considering length reduced
with a length reduction of about -50\% on Tedlium 3. Systems A and B are conventional ASR systems trained as joint CTC-attention models~\cite{kim2017joint} from scratch. Note that system A also uses a transformer-based LM when decoding. The baseline system E is our (approximately) reproduced version of system C~\cite{pan2023cosmic}; we assume that the WER gap between C and E are due to different layer types (4 transformer blocks vs.~12 E-Branchformer~\cite{kim2023branchformer}). Our DSU+LLM approach (G, H, J) demonstrates significantly better performance on all ASR test sets. Note that WavLM uses the VoxPopuli dataset in pre-training, so it is a partially seen domain for the DSU model, but an unseen domain for the speech adapter and LLM.

Both SLUE-VC and Fleurs are completely unseen domains for WavLM and the speech adapter. However, our model shows significantly better performance on these domains than the CTC compression baseline. This indicates that although the LLM is robust across text domains, the performance suffers when the input speech is out-of-domain for the encoder.  DSU may serve as a more general-purpose input, which abstracts away certain domain details, and so are beneficial for unseen speech domains in the real world.
The extremely low BLEU-1 scores on SLUE-SQA5 show that the model is unable to follow the instructions for a new task.
\vspace{-0.1cm}

\subsubsection{Adding SLU tasks}
\vspace{-0.1cm}
Next we use systems E, G, J from Table~\ref{tab:exp_asr}, and train them by adding more tasks.  We train from scratch when adding more data. The results are shown in Table~\ref{tab:exp_slu}.  We first add the SQA task (E1, G1, J1) and then add SA and NER with more ASR data (E2, G2, J2). 

Note that the SLUE-VC and SLUE-VP datasets are relatively small (\textless15h each) and are intended for low-resource conditions. At the same time, the tasks are also much simpler than SQA: SA is a classification task, and NER is a detection+classification task.
When we add these small SLU and ASR datasets in training, we observe a performance degradation in the CTC compression-based system for both ASR and SQA tasks (E1 vs.~E2 in Table~\ref{tab:exp_slu}). 
This indicates that useful context information may be discarded in the CTC-compressed sequence since it is optimized for ASR.
% We assume that this is because the CTC-compressed sequence is optimized for ASR, which may result in other useful context information being discarded.
In contrast, the DSU-based approach shows better or similar %range of 
performance when adding more datasets (G1 vs.~G2, and J1 vs.~J2).
For NER and SA tasks, 
%we tried to evaluate the performance, but 
we find that the model is not able to generate outputs that are well-formed enough to evaluate.
%the result enough to evaluate. 
We assume this is because the training sets are too small and the generative model is 
%being 
weak for classification tasks.

An interesting finding, shown in Table~\ref{tab:exp_s2tt}, is that the small amount of SLU training data appears to have unlocked the zero-shot capability of the model for speech-to-text translation (S2TT).  The model is able to follow the translation instruction and generate text in the target language without being trained on the Fleurs dataset or any translation task (E1 vs.~E2 and J1 vs.~J2).
As an ``oracle" reference, System K2 uses a text LLM instruction-tuned by feeding human transcriptions and the same S2TT text instruction in Table~\ref{tab:task_prompt}. 
This result suggests that we could expect an even more general model if we add more tasks and datasets in training the DSU-based model.

\subsubsection{Comparison of DSU types}
\vspace{-0.1cm}
Table~\ref{tab:exp_dsu} shows the results of varying the layer from 15 to 1 for extracting DSU. We also consider using MFCCs instead of pre-trained WavLM. When the embedding is extracted from a shallow layer, the ASR performance declines significantly. However, the BLEU score for the SQA task remains in a similar range. MFCC input causes only about 13\% degradation in BLEU compared to WavLM/21 (J2 vs.~L1). Training on ASR data is beneficial for the SQA task for DSU (J2 vs. J7), but not for MFCC (L1 and L2). We believe the model simply cannot generate the transcription given the ASR instruction, but understands the content enough to answer the given question using the SQA data.
In addition, neither the DSU-based approach (J7) nor the speech encoder using CTC (E3) shows any strength on the SQA task compared to MFCC (L2). This suggests that a better alternative audio quantization/tokenization could fill the gap between MFCC and WavLM/21 for universal speech tasks. 
Overall, this is a very intriguing result since %we 
it is typically assumed that the language model first transcribes speech, then understands it, and finally generates the answer. We will conduct further studies to determine whether the model does not need to transcribe the speech in order to answer, or instead the model just leverages pre-trained knowledge to answer, regardless of the audio document.
\vspace{-0.1cm}

%% file: sections/conclusion.tex
\vspace{-0.1cm}
Our study introduce the use of DSU to prompt LLMs with speech, 
%which has proven 
and has found it effective in both seen and unseen speech domains, even on a zero-shot task.
We demonstrate that ASR is not necessary for reasoning (SQA) if the speech representation is close to the raw signal. 
While our approach uses smaller-scale datasets %not as large-scale as 
than those in other work~\cite{wang2023slm} and does not rely on large-scale label generation with a teacher LLM~\cite{pan2023cosmic,tang2023salmonn,fathullah2023towards}, it is still %efficient enough 
able to train the LLM to follow instructions. 
In addition, all data and labels are publicly available, so this study should be easily reproducible. However, we note that we did not validate open-ended SQA performance, which we leave to future work.  Furthermore, the training process is inherently 2-step since the k-means clustering part is not differentiable. Future work could include using a neural audio codec or quantization method that enables end-to-end training with discrete units.

% In this study, we have introduced the use of discrete speech units based for prompting LLMs with speech.  We have shown the effectiveness of this approach, including on both seen and unseen speech domains, on a zero-shot task. 
% Particularly, ASR task is not necessarily needed for reasoning SQA if speech representation is close to raw signal. 
% % using DSUs extracted from a shallow layer.
% The datasets used in this study are not as large-scale as those in other recent LLM-based speech processing approaches~\cite{wang2023slm}, and we also do not rely on large-scale label generation with a teacher LLM~\cite{pan2023cosmic,tang2023salmonn,fathullah2023towards}, but the approach is still efficient enough to train the LLM to follow instructions. In addition, all data and labels are publicly available, so this study should be easily reproducible. However, we note that we did not validate open-ended SQA performance, which we leave to future work.  In addition, the training process is inherently 2-step since the k-means clustering part is not differentiable. Future work could include using a neural audio codec or quantization method that enables end-to-end training with discrete units. 

%% file: template.bbl
\begin{thebibliography}{10}
\providecommand{\url}[1]{#1}
\csname url@samestyle\endcsname
\providecommand{\newblock}{\relax}
\providecommand{\bibinfo}[2]{#2}
\providecommand{\BIBentrySTDinterwordspacing}{\spaceskip=0pt\relax}
\providecommand{\BIBentryALTinterwordstretchfactor}{4}
\providecommand{\BIBentryALTinterwordspacing}{\spaceskip=\fontdimen2\font plus
\BIBentryALTinterwordstretchfactor\fontdimen3\font minus \fontdimen4\font\relax}
\providecommand{\BIBforeignlanguage}[2]{{%
\expandafter\ifx\csname l@#1\endcsname\relax
\typeout{** WARNING: IEEEtran.bst: No hyphenation pattern has been}%
\typeout{** loaded for the language `#1'. Using the pattern for}%
\typeout{** the default language instead.}%
\else
\language=\csname l@#1\endcsname
\fi
#2}}
\providecommand{\BIBdecl}{\relax}
\BIBdecl

\bibitem{pan2023cosmic}
J.~Pan, J.~Wu, Y.~Gaur, S.~Sivasankaran, Z.~Chen, S.~Liu, and J.~Li, ``Cosmic: Data efficient instruction-tuning for speech in-context learning,'' \emph{arXiv preprint arXiv:2311.02248}, 2023.

\bibitem{wang2023slm}
M.~Wang, W.~Han, I.~Shafran, Z.~Wu, C.-C. Chiu, Y.~Cao, N.~Chen, Y.~Zhang, H.~Soltau, P.~K. Rubenstein \emph{et~al.}, ``Slm: Bridge the thin gap between speech and text foundation models,'' in \emph{{IEEE} Auto-matic Speech Recognition and Understanding Workshop (ASRU)}, 2023, pp. 1--8.

\bibitem{tang2023salmonn}
C.~Tang, W.~Yu, G.~Sun, X.~Chen, T.~Tan, W.~Li, L.~Lu, Z.~Ma, and C.~Zhang, ``Salmonn: Towards generic hearing abilities for large language models,'' \emph{arXiv preprint arXiv:2310.13289}, 2023.

\bibitem{fathullah2023towards}
Y.~Fathullah, C.~Wu, E.~Lakomkin, J.~Jia, Y.~Shangguan, J.~Mahadeokar, O.~Kalinli, C.~Fuegen, and M.~Seltzer, ``Towards general-purpose speech abilities for large language models using unpaired data,'' \emph{arXiv preprint arXiv:2311.06753}, 2023.

\bibitem{chu2023qwen}
Y.~Chu, J.~Xu, X.~Zhou, Q.~Yang, S.~Zhang, Z.~Yan, C.~Zhou, and J.~Zhou, ``Qwen-audio: Advancing universal audio understanding via unified large-scale audio-language models,'' \emph{arXiv preprint arXiv:2311.07919}, 2023.

\bibitem{hsu2021hubert}
W.-N. Hsu, B.~Bolte, Y.-H.~H. Tsai, K.~Lakhotia, R.~Salakhutdinov, and A.~Mohamed, ``{HuBERT: Self-Supervised Speech Representation Learning by Masked Prediction of Hidden Units},'' \emph{arXiv preprint arXiv:2106.07447}, 2021.

\bibitem{chen2022wavlm}
S.~Chen, C.~Wang, Z.~Chen, Y.~Wu, S.~Liu, Z.~Chen, J.~Li, N.~Kanda, T.~Yoshioka, X.~Xiao \emph{et~al.}, ``Wavlm: Large-scale self-supervised pre-training for full stack speech processing,'' \emph{IEEE Journal of Selected Topics in Signal Processing}, vol.~16, no.~6, pp. 1505--1518, 2022.

\bibitem{chang2023exploring}
X.~Chang, B.~Yan, K.~Choi, J.~Jung, Y.~Lu, S.~Maiti, R.~Sharma, J.~Shi, J.~Tian, S.~Watanabe \emph{et~al.}, ``Exploring speech recognition, translation, and understanding with discrete speech units: A comparative study,'' \emph{arXiv preprint arXiv:2309.15800}, 2023.

\bibitem{wu2023wav2seq}
F.~Wu, K.~Kim, S.~Watanabe, K.~J. Han, R.~McDonald, K.~Q. Weinberger, and Y.~Artzi, ``Wav2seq: Pre-training speech-to-text encoder-decoder models using pseudo languages,'' in \emph{IEEE ICASSP}, 2023, pp. 1--5.

\bibitem{lakhotia2021generative}
K.~Lakhotia, E.~Kharitonov, W.-N. Hsu, Y.~Adi, A.~Polyak, B.~Bolte, T.-A. Nguyen, J.~Copet, A.~Baevski, A.~Mohamed \emph{et~al.}, ``On generative spoken language modeling from raw audio,'' \emph{Transactions of the Association for Computational Linguistics}, vol.~9, pp. 1336--1354, 2021.

\bibitem{lakomkin2023end}
E.~Lakomkin, C.~Wu, Y.~Fathullah, O.~Kalinli, M.~L. Seltzer, and C.~Fuegen, ``End-to-end speech recognition contextualization with large language models,'' \emph{arXiv preprint arXiv:2309.10917}, 2023.

\bibitem{wu2023decoder}
J.~Wu, Y.~Gaur, Z.~Chen, L.~Zhou, Y.~Zhu, T.~Wang, J.~Li, S.~Liu, B.~Ren, L.~Liu \emph{et~al.}, ``On decoder-only architecture for speech-to-text and large language model integration,'' in \emph{{IEEE} Auto-matic Speech Recognition and Understanding Workshop (ASRU)}, 2023.

\bibitem{fathullah2023prompting}
Y.~Fathullah, C.~Wu, E.~Lakomkin, J.~Jia, Y.~Shangguan, K.~Li, J.~Guo, W.~Xiong, J.~Mahadeokar, O.~Kalinli \emph{et~al.}, ``Prompting large language models with speech recognition abilities,'' \emph{arXiv preprint arXiv:2307.11795}, 2023.

\bibitem{yu2023connecting}
W.~Yu, C.~Tang, G.~Sun, X.~Chen, T.~Tan, W.~Li, L.~Lu, Z.~Ma, and C.~Zhang, ``Connecting speech encoder and large language model for asr,'' \emph{arXiv preprint arXiv:2309.13963}, 2023.

\bibitem{tsunoo2023decoder}
E.~Tsunoo, H.~Futami, Y.~Kashiwagi, S.~Arora, and S.~Watanabe, ``Decoder-only architecture for speech recognition with ctc prompts and text data augmentation,'' \emph{arXiv preprint arXiv:2309.08876}, 2023.

\bibitem{hono2023integration}
Y.~Hono, K.~Mitsuda, T.~Zhao, K.~Mitsui, T.~Wakatsuki, and K.~Sawada, ``An integration of pre-trained speech and language models for end-to-end speech recognition,'' \emph{arXiv preprint arXiv:2312.03668}, 2023.

\bibitem{chen2023salm}
Z.~Chen, H.~Huang, A.~Andrusenko, O.~Hrinchuk, K.~C. Puvvada, J.~Li, S.~Ghosh, J.~Balam, and B.~Ginsburg, ``{SALM}: Speech-augmented language model with in-context learning for speech recognition and translation,'' \emph{arXiv preprint arXiv:2310.09424}, 2023.

\bibitem{gong2023joint}
Y.~Gong, A.~H. Liu, H.~Luo, L.~Karlinsky, and J.~Glass, ``Joint audio and speech understanding,'' in \emph{{IEEE} Auto-matic Speech Recognition and Understanding Workshop (ASRU)}, 2023.

\bibitem{chou2023toward}
J.-C. Chou, C.-M. Chien, W.-N. Hsu, K.~Livescu, A.~Babu, A.~Conneau, A.~Baevski, and M.~Auli, ``Toward joint language modeling for speech units and text,'' \emph{arXiv preprint arXiv:2310.08715}, 2023.

\bibitem{nguyen2024spirit}
T.~A. Nguyen, B.~Muller, B.~Yu, M.~R. Costa-Jussa, M.~Elbayad, S.~Popuri, P.-A. Duquenne, R.~Algayres, R.~Mavlyutov, I.~Gat \emph{et~al.}, ``Spirit-lm: Interleaved spoken and written language model,'' \emph{arXiv preprint arXiv:2402.05755}, 2024.

\bibitem{rubenstein2023audiopalm}
P.~K. Rubenstein, C.~Asawaroengchai, D.~D. Nguyen, A.~Bapna, Z.~Borsos, F.~d.~C. Quitry, P.~Chen, D.~E. Badawy, W.~Han, E.~Kharitonov \emph{et~al.}, ``Audiopalm: A large language model that can speak and listen,'' \emph{arXiv preprint arXiv:2306.12925}, 2023.

\bibitem{chung2021w2v}
Y.-A. Chung, Y.~Zhang, W.~Han, C.-C. Chiu, J.~Qin, R.~Pang, and Y.~Wu, ``W2v-bert: Combining contrastive learning and masked language modeling for self-supervised speech pre-training,'' in \emph{{IEEE} Auto-matic Speech Recognition and Understanding Workshop (ASRU)}.

\bibitem{zhang2023google}
Y.~Zhang, W.~Han, J.~Qin, Y.~Wang, A.~Bapna, Z.~Chen, N.~Chen, B.~Li, V.~Axelrod, G.~Wang \emph{et~al.}, ``Google usm: Scaling automatic speech recognition beyond 100 languages,'' \emph{arXiv preprint arXiv:2303.01037}, 2023.

\bibitem{pasad2021layer}
A.~Pasad, J.-C. Chou, and K.~Livescu, ``Layer-wise analysis of a self-supervised speech representation model,'' in \emph{{IEEE} Auto-matic Speech Recognition and Understanding Workshop (ASRU)}, 2021, pp. 914--921.

\bibitem{maiti2023voxtlm}
S.~Maiti, Y.~Peng, S.~Choi, J.-w. Jung, X.~Chang, and S.~Watanabe, ``Voxtlm: unified decoder-only models for consolidating speech recognition/synthesis and speech/text continuation tasks,'' \emph{arXiv preprint arXiv:2309.07937}, 2023.

\bibitem{zhang2023speechgpt}
D.~Zhang, S.~Li, X.~Zhang, J.~Zhan, P.~Wang, Y.~Zhou, and X.~Qiu, ``Speechgpt: Empowering large language models with intrinsic cross-modal conversational abilities,'' \emph{arXiv preprint arXiv:2305.11000}, 2023.

\bibitem{hernandez2018ted}
F.~Hernandez, V.~Nguyen, S.~Ghannay, N.~Tomashenko, and Y.~Est{\`{e}}ve, ``{TED-LIUM 3: Twice as Much Data and Corpus Repartition for Experiments on Speaker Adaptation},'' in \emph{Lecture Notes in Computer Science}, 2018.

\bibitem{shon2022slue}
S.~Shon, A.~Pasad, F.~Wu, P.~Brusco, Y.~Artzi, K.~Livescu, and K.~J. Han, ``Slue: New benchmark tasks for spoken language understanding evaluation on natural speech,'' in \emph{{IEEE} ICASSP}, 2022, pp. 7927--7931.

\bibitem{conneau2023fleurs}
A.~Conneau, M.~Ma, S.~Khanuja, Y.~Zhang, V.~Axelrod, S.~Dalmia, J.~Riesa, C.~Rivera, and A.~Bapna, ``Fleurs: Few-shot learning evaluation of universal representations of speech,'' in \emph{IEEE Spoken Language Technology Workshop (SLT)}, 2022, pp. 798--805.

\bibitem{shon2022slue2}
S.~Shon, S.~Arora, C.-J. Lin, A.~Pasad, F.~Wu, R.~Sharma, W.-L. Wu, H.-Y. Lee, K.~Livescu, and S.~Watanabe, ``Slue phase-2: A benchmark suite of diverse spoken language understanding tasks,'' \emph{arXiv preprint arXiv:2212.10525}, 2022.

\bibitem{hu2021lora}
E.~J. Hu, Y.~Shen, P.~Wallis, Z.~Allen-Zhu, Y.~Li, S.~Wang, L.~Wang, and W.~Chen, ``Lora: Low-rank adaptation of large language models,'' \emph{arXiv preprint arXiv:2106.09685}, 2021.

\bibitem{jiang2023mistral}
A.~Q. Jiang, A.~Sablayrolles, A.~Mensch, C.~Bamford, D.~S. Chaplot, D.~d.~l. Casas, F.~Bressand, G.~Lengyel, G.~Lample, L.~Saulnier \emph{et~al.}, ``Mistral 7b,'' \emph{arXiv preprint arXiv:2310.06825}, 2023.

\bibitem{chang2023exploration}
X.~Chang, B.~Yan, Y.~Fujita, T.~Maekaku, and S.~Watanabe, ``Exploration of efficient end-to-end asr using discretized input from self-supervised learning,'' \emph{arXiv preprint arXiv:2305.18108}, 2023.

\bibitem{kim2017joint}
S.~Kim, T.~Hori, and S.~Watanabe, ``{Joint {CTC}-attention based end-to-end speech recognition using multi-task learning},'' in \emph{{IEEE} ICASSP}, 2017.

\bibitem{kim2023branchformer}
K.~Kim, F.~Wu, Y.~Peng, J.~Pan, P.~Sridhar, K.~J. Han, and S.~Watanabe, ``E-branchformer: Branchformer with enhanced merging for speech recognition,'' in \emph{IEEE Spoken Language Technology Workshop (SLT)}, 2022, pp. 84--91.

\end{thebibliography}
